\documentclass[runningheads]{llncs}
\usepackage{graphicx}
\usepackage{amsmath,amssymb} 
\usepackage{color}
\usepackage[width=122mm,left=12mm,paperwidth=146mm,height=193mm,top=12mm,paperheight=217mm]{geometry}
\usepackage[sort,nocompress]{cite}
\begin{document}
\pagestyle{headings}
\mainmatter

\title{Gated Siamese Convolutional Neural Network Architecture for Human Re-Identification} 

\titlerunning{Gated S-CNN Architecture for Human Re-Identification}

\authorrunning{Rahul Rama Varior, Mrinal Haloi, and Gang Wang}

\author{Rahul Rama Varior, Mrinal Haloi, and Gang Wang\thanks{Corresponding author.}}


\institute{School of Electrical and Electronic Engineering, Nanyang Technological University\\
	\email{ \{rahul004,mhaloi,wanggang\}@ntu.edu.sg}
}

\maketitle

\begin{abstract}
Matching pedestrians across multiple camera views, known as human re-identification, is a challenging research problem that has numerous applications in visual surveillance. With the resurgence of Convolutional Neural Networks (CNNs), several end-to-end deep Siamese CNN architectures have been proposed for human re-identification with the objective of projecting the images of similar pairs (i.e. same identity) to be closer to each other and those of dissimilar pairs to be distant from each other. However, current networks extract fixed representations for each image regardless of other images which are paired with it and the comparison with other images is done only at the final level. In this setting, the network is at risk of failing to extract finer local patterns that may be essential to distinguish positive pairs from hard negative pairs. In this paper, we propose a gating function to selectively emphasize such fine common local patterns by comparing the mid-level features across pairs of images. This produces flexible representations for the same image according to the images they are paired with. We conduct experiments on the CUHK03, Market-1501 and VIPeR datasets and demonstrate improved performance compared to a baseline Siamese CNN architecture. 
\end{abstract}

\section{Introduction}
\label{sec:intro}

Matching pedestrians across multiple camera views, also known as human re-identification, is a research problem that has numerous potential applications in visual surveillance. The goal of the human re-identification system is to retrieve a set of images captured by different cameras (gallery set) for a given query image (probe set) from a certain camera. Human re-identification is a very challenging task due to the variations in illumination, pose and visual appearance across different camera views. With the resurgence of Convolutional Neural Networks (CNNs), several deep learning methods \cite{yi2014deep,ejazdeep2015,cuhk03} were proposed for human re-identification. Most of the frameworks are designed in a siamese fashion that integrates the tasks of feature extraction and metric learning into a single framework.

\begin{figure}[!t]
\centering
\includegraphics[width=0.9\linewidth]{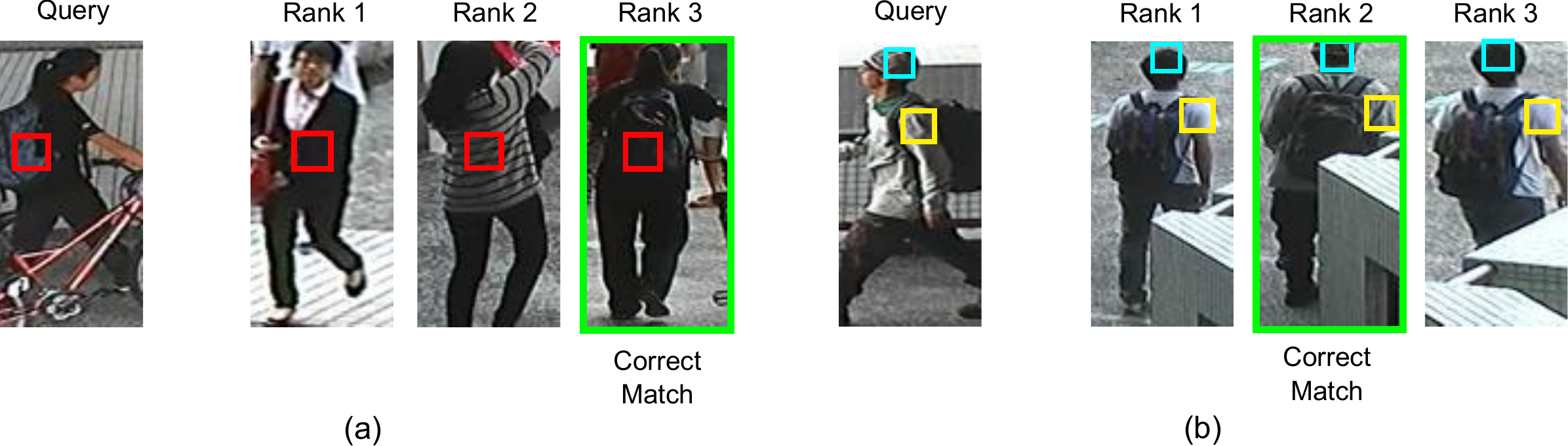}
\caption{{\bf Example case: }Results obtained using a S-CNN. Red, Blue and Yellow boxes indicate some sample corresponding patches extracted from the images along the same horizontal row. See text for more details. {\bf Best viewed in color}}
\label{fig:example}
\end{figure}

The central idea behind a Siamese Convolutional Neural Network (S-CNN) is to learn an embedding where similar pairs (i.e. images belonging to the same identity) are close to each other and dissimlar pairs (i.e. images belonging to different identities) are separated by a distance defined by a parameter called `margin'. In this paper, we first propose a baseline S-CNN architecture that can outperform majority of the deep learning architectures as well as other handcrafted approaches for human re-identification on challenging human re-identification datasets, the CUHK03 \cite{cuhk03}, the Market-1501 \cite{market} and the VIPeR \cite{viper} dataset. 

The major drawback of the S-CNN architecture is that it extract fixed representations for each image without the knowledge of the paired image. This setting results in a risk of failing to capture and propagate the local patterns that are necessary to increase the confidence level (i.e., reducing the distances) in identifying the correct matches. Figure \ref{fig:example} (a) and (b) shows two queries and the retrieved matches at the top 3 ranks using a S-CNN architecture. Even though there are obvious dissimilarities among the top 3 matches for a human observer in both the cases, the network fails to identify the correct match at Rank 1. For example, the patches corresponding to the `bag' (indicated by red boxes) in Figure \ref{fig:example} (a) and the patches corresponding to the `hat' (indicated by blue boxes) in Figure \ref{fig:example} (b) could be helpful to distinguish between the top retrieved match and the actual positive pairs. However, a network that fails to capture and propagate such finer details may not perform well in efficiently distinguishing positives from hard-negatives.

CNNs extract low-level features at the bottom layers and learn more abstract concepts such as the parts or more complicated texture patterns at the mid-level. Since the mid-level features are more informative compared to the higher-level features, the finer details that may be necessary to increase the similarity for positive pairs can be more evident at the middle layers. Hence, we propose a gating function to compare the extracted local patterns for an image pair starting from the mid-level and promote (i.e. to amplify) the local similarities along the higher layers so that the network propagates more relevant features to the higher layers of the network. Additionally, during training phase, the mechanisms inside the gating function also boost the back propagated gradients corresponding to the amplified local similarities. This encourages the lower and middle layers to learn filters to extract more locally similar patterns that discriminate positive pairs from negative pairs. Hereafter, we refer to the proposed gating function as `the Matching Gate' (MG).

The primary challenge in developing the matching gate is that it should be able to compare the local features across two views effectively and select the common patterns. Due to pose change across two views, features appearing at one location may not necessarily appear in the same location for its paired image.  Since all the images are resized to a fixed scale, it is reasonable to assume a horizontal row-wise correspondence. Therefore, the matching gate first summarizes the features along each horizontal stripe for a pair of images and compares it by taking the Euclidean distance along each dimension of the obtained feature map. Once the distances between each individual dimensions are obtained, a Gaussian activation function is used to output a similarity score ranging from $0 - 1$ where $0$ indicates that the stripe features are dissimilar and $1$ indicating that the stripe features are similar. These values are used to gate the stripe features and finally, the gated features are added to the input features to boost them thus giving more emphasis to the local similarities across view-points. Our approach does not require any part-level correspondence annotation between image pairs during the training phase as it directly compares the extracted mid-level features along corresponding horizontal stripes. Additionally, the proposed matching gate is formulated as a differentiable parametric function to facilitate the end-to-end learning strategy of typical deep learning architectures. To summarize, the major contributions of the proposed work are:

\begin{itemize}
\item We propose a baseline siamese convolutional neural network architecture that can outperform majority of the existing deep learning frameworks for human re-identification.
\item To incorporate run time feature selection and boosting into the S-CNN architecture, we propose a novel matching gate that can boost the common local features across two views. This encourages the network to learn filters that can extract subtle patterns to discriminate hard-negatives from positive pairs. The proposed matching gate is differentiable to facilitate end-to-end training of the S-CNN architecture.
\item We conduct experiments on the CUHK03 \cite{cuhk03}, Market-1501 \cite{market} and the VIPeR \cite{viper} datasets for human re-identification and prove the effectiveness of our approach. The proposed framework also achieves promising results compared to the state-of-the-art algorithms.
\end{itemize}

\section{Related Works}

\subsection{Human Re-Identification}
Existing research on human re-identification mainly concentrates on two aspects: (1) Developing a new feature representation \cite{lomo,varior2016learning,yangcolor2014,salientcolorECCV14,zhang2014novel,bicovma2012,custompict,kviatkovsky2013color} and (2) Learning a distance metric \cite{lomo,li2012human,pedagadilfda,ZhenliShiyu_CVPR2013,eccv14prid,variorMarginalization,mlapg,Su_2015_ICCV}. Novel feature representations were proposed \cite{lomo,varior2016learning,bicovma2012} to address the challenges such as variations in illumination, pose and view-point. Scale Invariant Feature Transforms \cite{lowe2004distinctive,zhao2013unsupervised,zhao2013person}, Scale Invariant Local Ternary Patterns \cite{lomo,siltp}, Local Binary Patterns \cite{eccv14prid,lbp}, Color Histograms \cite{lomo,eccv14prid,zhao2013unsupervised,zhao2013person} or Color Names \cite{market,salientcolorECCV14} etc. are the basis of the majority of these feature representations developed for human re-identification. Several Metric Learning algorithms such as Locally adaptive Decision Functions (LADF) \cite{ZhenliShiyu_CVPR2013}, Cross-view Quadratic Discriminant Analysis (XQDA) \cite{lomo}, Metric Learning with Accelerated Proximal Gradient (MLAPG) \cite{mlapg}, Local Fisher Discriminant Analysis (LFDA) \cite{pedagadilfda} and its kernel variant (k-LFDA) \cite{eccv14prid} were proposed for human re-identification achieving remarkable performance in several benchmark datasets. However, different from all the above works, our approach is modeled based on the Siamese Convolutional Neural Networks (S-CNN) \cite{siamese,drlim} that can learn an embedding where similar instances are closer to each other and dissimilar images are distant from each other from raw pixel values. 

\subsubsection{Deep Learning for Human Re-Identification: }Convolutional Neural Networks have achieved phenomenal results on several computer vision tasks \cite{resnet,inception,vggnet,shuaiconv}. In the recent years, several CNN architectures \cite{ejazdeep2015,mcpbc,li2014deepreid,slstm,sicir,domainguided,yi2014deep} have been proposed for human re-identification. The first Siamese CNN (S-CNN) architecture for human re-identification was proposed in \cite{yi2014deep}. The system (DML) consists of a set of 3 S-CNNs for different regions of the image and the features are combined by using a cosine similarity as the connection function. Finally a binomial deviance is used as the cost function to optimize the network end-to-end. Local body-part based features and the global features were modeled using a Multi-Channel CNN framework in \cite{mcpbc}. Deep Filter Pairing Neural Network (FPNN) was introduced in \cite{cuhk03} to jointly handle misalignment, photometric and geometric transformations, occlusion and cluttered background. In \cite{ejazdeep2015}, a cross-input neighborhood difference module was proposed to extract the cross-view relationships of the features and have achieved impressive results in several benchmark datasets. A recent work \cite{sicir} also attempts to model the cross-view relationships by jointly learning subnetworks to extract the single image as well as the cross image representations. In \cite{domainguided}, domain guided dropout was introduced for selecting the appropriate neuron for the images belonging to a given domain.  A Long-Short Term Memory (LSTM) based architecture was proposed in \cite{slstm} to model the contextual dependencies and selecting the relevant contexts to improve the discriminative capabilities of the local features. Different from all the above works, the proposed matching gate aims at comparing features at multiple levels (different layers) to boost the local similarities and enhance the discriminative capability of the propagated local features. The proposed gating function is flexible (in architecture) and differentiable to facilitate end-to-end learning strategy of deep neural networks.

\subsection{Gating Functions}
Gating functions have been proven to be an important component in deep neural networks \cite{highway,lstm}. Gating mechanisms such as the input gates and output gates were proposed in Long-Short Term Memory (LSTM) \cite{lstm} cells for regulating the information flow through the network. Further, LSTM unit with forget gate \cite{forgetgate} was proposed to reset the internal states based on the inputs. Inspired by the LSTM, Highway Networks \cite{highway} were proposed to train very deep neural networks by introducing gating functions into the CNN architecture. More recently, `Trust Gates' were introduced in \cite{juneccv16} to handle the noise and occlusion in 3D skeleton data for action recognition. However, the proposed matching gate is modeled entirely in a different context in terms of its architecture and purpose; i.e., the goal of the matching gate is to compare the local feature similarities of input pairs from the mid-level through the higher layers and weigh the common local patterns based on the similarity scores. This will enable the lower layers of the network to learn filters that can discriminate the local patterns of positive pairs from negative pairs. Additionally, to the best of our knowledge, the proposed work is the first of its nature to introduce differentiable gating functions in siamese architecture for human re-identification.

\section{Proposed Model}
\label{sec:model}
In this section, we first describe our baseline S-CNN architecture and further introduce the Matching Gate to address the limitations of the baseline S-CNN architecture.

\subsection{Model Architecture}
\begin{table}[!t]

	\small
	\renewcommand{\arraystretch}{1}
	\setlength{\tabcolsep}{1.5pt}
	\caption{Proposed Baseline Siamese Convolutional Neural Network architecture.}
	\label{table:archi}
	\centering
	\scalebox{0.6}{
	\begin{tabular}{|c|c|c|c|c|c|c|c|c|c|c|c|}
		\hline
		 {\bf \begin{tabular}{@{}c@{}} Input \end{tabular}} & {\bf \begin{tabular}{@{}c@{}} Conv\\Block - P2 \end{tabular}} & {\bf \begin{tabular}{@{}c@{}} Max\\Pool\end{tabular}} & {\bf \begin{tabular}{@{}c@{}} Conv\\Block - P1  \end{tabular}} & {\bf \begin{tabular}{@{}c@{}} Max\\Pool\end{tabular}}  & {\bf \begin{tabular}{@{}c@{}} Conv\\Block - P1  \end{tabular}} & {\bf \begin{tabular}{@{}c@{}} Max\\Pool\end{tabular}}  & {\bf \begin{tabular}{@{}c@{}} Conv\\Block \end{tabular}}  & {\bf \begin{tabular}{@{}c@{}} Conv\\Block \end{tabular}}  & {\bf \begin{tabular}{@{}c@{}} Conv\\Block \end{tabular}}  & {\bf \begin{tabular}{@{}c@{}} Conv\\Block \end{tabular}} \\
		 \hline
		 $128 \times 64$ & $5\times5\times3\times32$ & $2\times 2$ & $3\times3\times32\times50$ & $2\times 2$ & $3\times3\times50\times32$ & $2\times 2$ &$1\times4\times32\times32$ & $1\times3\times32\times32$ & $1\times3\times32\times32$ & $16\times1\times32\times150$ \\  
		 \hline
		 
	\end{tabular}	}
	{\tiny ConvBlock - Convolution -> Batch Normalization -> Parametric Rectified Linear Unit} \\
	{\tiny P2 and P1 - zero padding the input with 2 pixels and 1 pixel on all sides respectively before convolution}
\end{table}
\subsubsection{Baseline Siamese CNN Architecture: }The fundamental CNN architecture is modeled in a siamese fashion optimized by the contrastive loss function proposed in \cite{drlim}. Table \ref{table:archi} summarizes the proposed Siamese CNN architecture. All the inputs are resized to a resolution of $128 \times 64$ and the mean image computed on the training set is subtracted from all the images. The description of the proposed S-CNN layers is as follows. First, we limit the number of pooling layers to only $3$ so that it results in less information loss as the features propagate through the network. Second, we also use asymmetric filtering in layers $4-6$ to preserve the number of rows at the output of the third layer while reducing the number of `columns' progressively to 1. This strategy is inspired by the technique introduced in \cite{lomo} in which the features along a single row is pooled to make the final feature map to a shape (number of rows) $\times 1$. It also helps to reduce the number of parameters compared to symmetric filters. Further, this feature map is fed into a fully connected layer which is the last layer of our network. Finally, we also incorporate some of the established state-of-the-art techniques to the proposed S-CNN architecture. As suggested in VGG-Net \cite{vggnet}, we use smaller convolutional filters to reduce the number of parameters to be learned while making the framework deeper. We also employ Batch Normalization \cite{bnorm} for standardizing the distribution of the inputs to each layer which helps in accelerating the training procedure. Parametric rectified linear unit (PReLU) \cite{prelu} was used as the non-linear activation function as it has shown better convergence properties and performance gains with little risk of over-fitting. More results and analysis about the design choices are given in the supplementary material. The proposed S-CNN architecture outperforms majority of the existing approaches for human re-identification. However, as discussed in Section \ref{sec:intro}, the S-CNN model is not capable of adaptively emphasizing the local features that may be helpful to distinguish the correct matches from hard-negative pairs during run time. Therefore, we propose a matching gate to address this drawback. Below we give the details of the proposed module.

\subsubsection{Matching Gate: }

\begin{figure}[!t]
\centering
\includegraphics[width=0.9\linewidth]{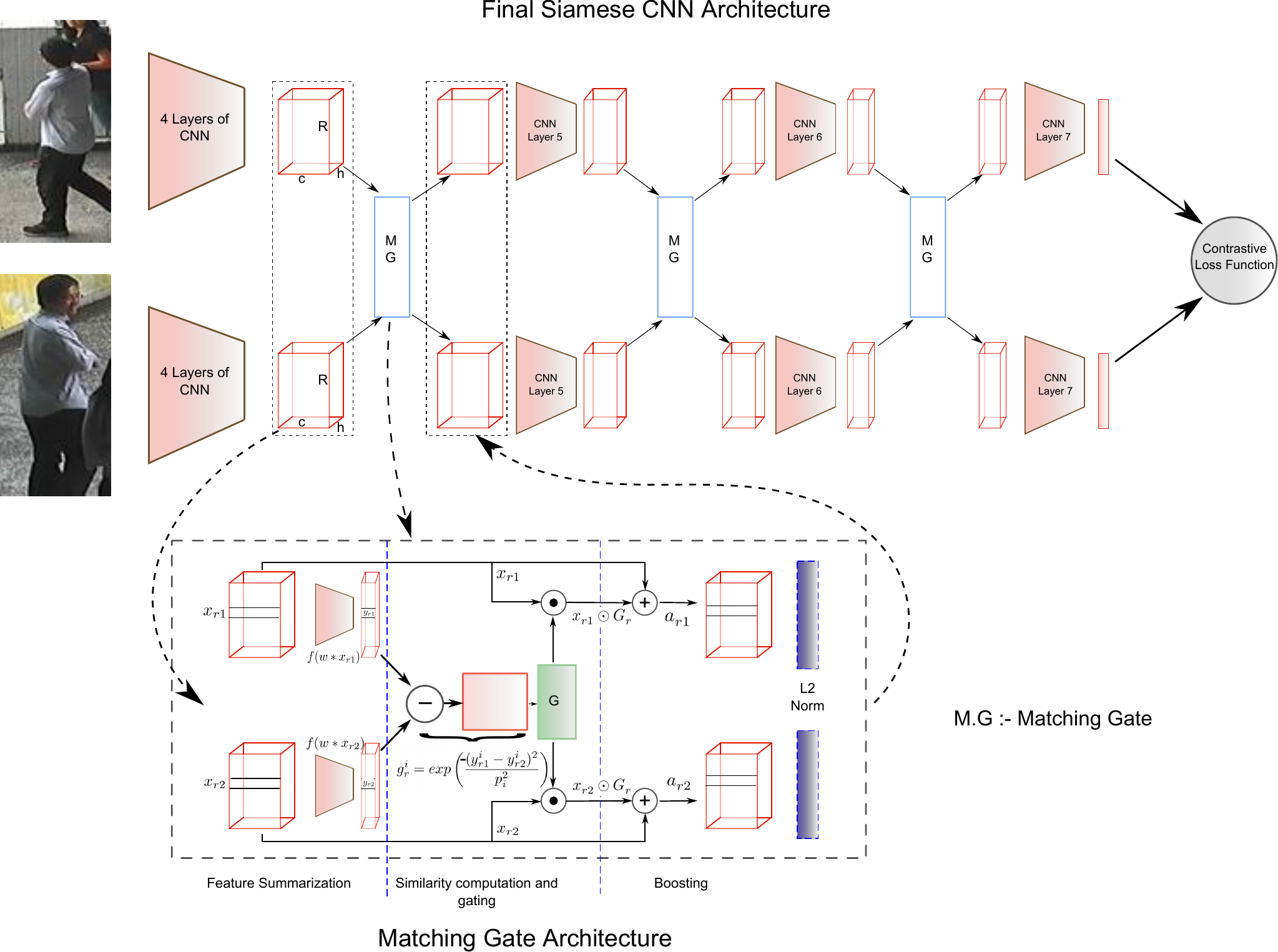}
\caption{{\bf Proposed architecture: }The proposed architecture is a modified version of our baseline S-CNN proposed in Table \ref{table:archi}. The matching gate is inserted between layers $4-5$, $5-6$ and $6-7$. The detailed architecture of the gating function is also shown in the figure. See text for details. {\bf Best viewed in color}}
\label{fig:model}
\end{figure}

The proposed matching gate (MG) receives input activations from the previous convolutional block, compares the local features along a horizontal stripe and outputs a gating mask indicating how much more emphasis should be paid to each of the local patterns. Figure \ref{fig:model} illustrates the proposed final architecture with the gating function. The various components of the proposed MG are given below.

\begin{enumerate}
\item {\bf Feature summarization: } The feature summarization unit aggregates the local features along a horizontal stripe in an image. This is necessary due to the pose changes of the pedestrian images across different views. For instance, as shown in Figure \ref{fig:example}, the local features (indicated by red, blue and yellow boxes) appearing in one view may not be exactly at the same position in the other view, but it is very likely to be along the same horizontal region. 

Let ${\bf x_{r1}} \in \mathbb{R}^{ 1 \times c \times h}$ be the input stripe features from the $r^{th}$ row of a feature map at the input of the MG from one view point and ${\bf x_{r2}}\in \mathbb{R}^{ 1 \times c \times h}$ be the corresponding input stripe features from the other view point. Here, $c$ denotes the number of columns and $h$ denotes the depth of the input feature map. Given ${\bf x_{r1}}$ and ${\bf x_{r2}}$, we propose to use a convolution strategy followed by the parametric rectified linear unit activation (PReLU) to summarize the features along the row resulting in feature vectors ${\bf y_{r1}}$ and ${\bf y_{r2}}$ respectively with dimensions $\mathbb{R}^{ 1 \times 1 \times h}$. The input features ${\bf x_{r1}}$ and ${\bf x_{r2}}$, are convolved with filters ${\bf w} \in \mathbb{R}^{1\times c\times h \times h}$ without any padding. This will compute the combination of different extracted patterns along each of the feature maps of ${\bf x_{r1}}$ and ${\bf x_{r2}}$.

Mathematically, it can be expressed as 
\begin{equation}
\label{eqn:conv}
{\bf y_{r1}} = f({\bf w} \ast {\bf x_{r1}}); \quad {\bf y_{r2}} = f({\bf w} \ast {\bf x_{r2}})
\end{equation}
where `$\ast$' denotes the convolution operation and $f(.)$ denotes the PReLU activation function. The bias is omitted in equation (\ref{eqn:conv}) for brevity. The parameters ${\bf w}$ and bias of the summarization unit can be learned along with the other parameters of the matching gate through back-propagation.

\item {\bf Feature Similarity computation: }
Once the features along a horizontal stripe are summarized across the two views, the similarity between them is computed. The similarity is computed by calculating the Euclidean distance along each dimension `$h$' of the summarized features. Computing the distance between each dimension is important as the gating function must have the flexibility to smoothly turn `on' or turn `off' each of the extracted patterns in the feature map. Once the distance is computed, a Gaussian activation function is used to obtain the gate values. The value of the Gaussian activation function varies from $0-1$ and acts as a smooth switch for the input features. It also helps the function to be differentiable which is essential for end-to-end training of the S-CNN framework. Mathematically the gating value for each of the dimensions along row `$r$' can be obtained as given below;
\begin{equation}
{\bf g_r}^i = exp\left(\frac{-({\bf y_{r1}}^i - {\bf y_{r2}}^i)^2}{{\bf p}_i^2}\right)
\end{equation}
where ${\bf g_r}^i, {\bf y_{r1}}^i$ and ${\bf y_{r2}}^i$ denotes the $i^{th}$ ($i=\{1,2,\dots,h\}$) dimension of the gate values (${\bf g_r}$), ${\bf y_{r1}}$ and ${\bf y_{r2}}$ respectively for the $r^{th}$ row. The parameter ${\bf p}_i$ decides the variance of the Gaussian function and the optimal value can be learned during the training phase. It is particularly important to set a higher initial value for ${\bf p}_i$ to ensure smooth flow of feature activations and gradients during forward and backward pass in the initial iterations of the training phase. Further, the network can decide the variance of the Gaussian function for each dimension by learning an optimal ${\bf p}_i$. 
\item {\bf Filtering and Boosting the features: } Once the gate values (${\bf g_r}$) are computed, each dimension along a row of the input is gated with the corresponding dimension of ${\bf g_r}$. The computed gate values will be of dimensions $ \mathbb{R}^{ 1 \times 1 \times h}$ and is repeated $c$ times horizontally to obtain $ {\bf G_r} \in \mathbb{R}^{ 1 \times c \times h}$ matrix and further an element wise product is computed with the input stripe features ${\bf x_{r1}}$ and ${\bf x_{r2}}$. This will `select' the common patterns along a row from the images appearing in both views. To boost these selected common patterns, the input is again added to these gated values. Mathematically, each dimension of the boosted output can be written as
\begin{eqnarray}
\label{boost}
{\bf a_{r1}}^i &=& {\bf x_{r1}}^i + {\bf x_{r1}}^i \odot {\bf G_r}^i \\
{\bf a_{r2}}^i &=& {\bf x_{r2}}^i + {\bf x_{r2}}^i \odot {\bf G_r}^i \\
{\bf G_r}^i &=& [{\bf g_r}^i, {\bf g_r}^i, \dotsc, {\bf g_r}^i]_{repeated\text{ }c\text{ }times}
\end{eqnarray}
 $\text{where } {\bf a_{r1}}^i,\text{ } {\bf a_{r2}}^i,\text{ } {\bf x_{r1}}^i,\text{ } {\bf x_{r2}}^i,\text{ } {\bf G_r}^i \in \mathbb{R}^{1\times c \times 1}$.
Once the boosted output ${\bf a_{r1}}$ and ${\bf a_{r2}}$ are obtained, we perform an $L2$ normalization across channels and the obtained features are propagated to the rest of the network. From Equations (\ref{boost}) and (4), we can understand that the gradients with respect to the `selected' ${\bf x_{r1}}$ and ${\bf x_{r2}}$ will also be boosted during the backward pass. This will encourage the lower layers of the network to learn filters that can extract patterns that are more similar for positive pairs.
\end{enumerate}

The key advantages of the proposed MG is that it is flexible in its architecture as well as differentiable. If the optimal variance factor ${\bf p}$ is learned to be high, it facilitates maximum information flow from the input to output and conversely if it is learned to be a low value, it allows only very similar patches to be boosted. The network learns to identify the optimal ${\bf p}$ for each dimension from the training data which results in a matching gate that is flexible in its functioning. Alongside learning an optimal ${\bf p}$, the network also learns the parameter ${\bf w}$ and the bias in Equation (\ref{eqn:conv}) to summarize the features along a horizontal stripe. Additionally, the MG can be inserted in between any layers or multiple layers in the network as it is a differentiable function. This will also facilitate end-to-end learning strategy in deep networks.

\subsubsection{Final Architecture: }The final architecture of the proposed system is shown in Figure \ref{fig:model}. The baseline network is designed in such a way as to reduce the width of the feature map progressively without reducing the height from layers $4-6$. This is essential to address the pose change of the human images across cameras while preserving the finer row-wise characteristics. As shown in the figure \ref{fig:model}, we inserted the proposed MG between the last $4$ layers once the number of rows of the propagated feature maps is fixed. 

\subsection{Training and Optimization}

\subsubsection{Input preparation: }Siamese networks take image pairs as inputs. Therefore, we first pair all the images in the training set with a label `$1$' indicating negative pairs and `$0$' indicating the positive pairs. For large datasets, the number of negative image pairs will be orders of magnitude higher than the number of positive pairs. To alleviate this bias in the training set, we perform artificial augmentation of the data by flipping the images and randomly translating them following \cite{ejazdeep2015}, to increase the number of positive pairs as well as sample approximately $5$ times the number of positive image pairs, as negative image pairs for each subject. The mean image computed from all the training images is subtracted from all the images and the input pairs are fed to the network.

\subsubsection{Training: } Both the baseline S-CNN model and the proposed architecture (Figure \ref{fig:model}) are trained from scratch in an end-to-end manner with a batch size of $100$ pairs in an iteration. The weight parameters (i.e. filters) of the networks are initialized uniformly following \cite{prelu}. The gradients with respect to the feature vectors at the last layer are computed from the contrastive loss function and back-propagated to the lower layers of the network. 
Once all the gradients are computed at all the layers, we use mini batch stochastic gradient descent (SGD) to update the parameters of the network. Specifically, we use the adaptive per-parameter update strategy called the RMSProp \cite{rmsprop} to update the weights. The decay parameter for RMSProp is fixed to $0.95$ following previous works \cite{karpathyweak} and the margin for the contrastive loss function is kept as $1$. Training is done for $20$ epochs with an early stopping strategy based on the saturation of the validation set performance. The initial learning rate is set to $0.002$ and reduced by a factor of $0.9$ after each epoch. The main hyper-parameter of the MG is the initial value of ${\bf p}$. We set this value to $4$ initially and the network discovers the optimal value during learning. More details on parameter tuning and validation are given in the supplementary material.

\subsubsection{Testing: }During testing, each query image has to be paired with all the gallery images and passed to the network. 
The Euclidean distance between the feature vectors obtained at the last layer is used to compare two input images. Once the distance between the query image and all the images in the gallery set are obtained, it is sorted in ascending order to find the top matches. The above procedure is done for all the query images and the final results are obtained. Finally, we also aggregate the matching scores over all epochs by averaging them to obtain the reported results. For an identity with multiple query images, the distances obtained for each query are rescaled in the range of $0-1 $ and then averaged.

\section{Experiments}
\label{sec:exp}
We provide a comprehensive evaluation of the proposed S-CNN architecture with the matching gate by comparing it against the baseline S-CNN architecture as well as other state-of-the-art algorithms for human re-identification. Majority of the human re-identification systems are evaluated based on the Cumulative Matching Characteristics by treating human re-identification as a ranking problem. However, in \cite{market}, human re-identification is treated as a retrieval problem and the mean average precision (mAP) is also reported along with the Rank - 1 accuracy (R1 Acc). For a fair comparison, we report both mAP as well as the performance at different ranks for CUHK03 dataset and mAP and R1 Acc for Market-1501 dataset. We also report both single-query (SQ) as well as multi-query (MQ) evaluation results for both of the above datasets. For VIPeR dataset, we report only the CMC as it is the relevant measure \cite{market}.
All the implementations are done in MATLAB-R2015b and we use the MatConvNet package \cite{matconvnet} for implementing all the proposed frameworks. Experiments were run on NVIDIA-Tesla K40 GPU and it took approximately 40-50 minutes per epoch on the CUHK03 dataset.

\subsection{Datasets and settings}Experiments were conducted on challenging benchmark datasets for human re-identification, the Market-1501 \cite{market} dataset, the CUHK03 \cite{cuhk03} dataset and the VIPeR \cite{viper} dataset. Below, we give the details of the datasets.

\subsubsection{Market-1501: }The Market-1501 dataset contains $32668$ annotated bounding boxes of 1501 subjects captured from $6$ cameras and is currently the largest dataset for human re-identification. The bounding boxes for the pedestrian images are obtained by using deformable parts model detectors. Therefore, the bounding boxes are not as ideal as the ones generated by human annotators and there are also several mis-detections which make the dataset very challenging. Following the standard evaluation protocols in \cite{market}, the dataset is split into $751$ identities for training and $750$ identities for testing. 

\subsubsection{CUHK03: }CUHK03 dataset contains $13164$ images of $1360$ subjects collected on the CUHK campus. Authors of \cite{cuhk03} provide two different settings for evaluating on this dataset, `detected' with automatically generated bounding boxes and `labeled' with human annotated bounding boxes. All the experiments presented in this paper follow the `detected' setting as this is closer to the real-world scenario. Following the splitting settings provided in \cite{cuhk03}, evaluation is conducted $20$ times with $100$ test subjects and the average result obtained at different ranks is reported. We also use $100$ identities from the training set for cross-validation leaving out $1160$ identities for training the network.

\subsubsection{VIPeR: } VIPeR dataset consists of $1264$ images belonging to $632$ subjects captured using $2$ cameras. The dataset is relatively small and the number of distinct identities as well as positive pairs per identity for training are very less compared to the other datasets. Therefore, we conduct data augmentation as well as transfer learning from Market-1501 and CUHK03 datasets. For transfer learning, we remove the last fully connected layer in our baseline S-CNN architecture and then fine-tune the network using the VIPeR dataset. Removing the last fully connected layer was to avoid over-fitting by reducing the number of parameters. For the gated S-CNN framework, the MGs are inserted between layers $4-5$ and $5-6$. Other experimental settings are kept the same as in \cite{ejazdeep2015}.

\begin{table}[!t]

	\scriptsize
	\renewcommand{\arraystretch}{1.1}
	\setlength{\tabcolsep}{2.5pt}
	\caption{Performance Comparison of state-of-the-art algorithms for the Market-1501 dataset. Proposed baseline S-CNN architecture outperforms the previous works for Market-1501 dataset. The S-CNN architecture with the gating function advances the state-of-the-art results on the Market-1501 dataset.}
	\label{market}
	\centering
	{
	\begin{tabular}{|c|c|c|c|c|c|}
		\hline
		\bfseries Method & \bfseries Rank 1 & \bfseries mAP 
		\\\hline\hline			
		SDALF \cite{sdalf} & 20.53 & 8.20 \\
		eSDC \cite{zhao2013unsupervised} & 33.54 & 13.54 \\
		BoW \cite{market} - (SQ)  & 34.40 & 14.09 \\

		DNS \cite{dns} - (SQ) & {61.02} & {35.68} \\		
		\hline
		{\bf \begin{tabular}{@{}c@{}} Ours - Baseline - S-CNN - (SQ) \end{tabular}} & {\bf 62.32} & {\bf 36.23} \\		
		{\bf \begin{tabular}{@{}c@{}} Ours - With Matching Gate - (SQ) \end{tabular}} &{\bf 65.88}	  & {\bf 39.55}  \\
		\hline		
		{\begin{tabular}{@{}c@{}} BoW \cite{market} - (MQ)\end{tabular}} & 42.14&19.20 \\
		{\begin{tabular}{@{}c@{}} BoW + HS \cite{market} - (MQ) \end{tabular}} &47.25 &21.88 \\		
		S-LSTM \cite{slstm} - (MQ) &{61.60}	  & {35.31} \\		
		DNS \cite{dns} - (MQ) & {71.56} & {46.03} \\		
		\hline

		{\bf \begin{tabular}{@{}c@{}} Ours - Baseline - S-CNN - (MQ) \end{tabular}} & {\bf 72.92} & {\bf 45.39} \\					
		{\bf \begin{tabular}{@{}c@{}} Ours - With Matching Gate - (MQ) \end{tabular}} &{\bf 76.04}	  & {\bf 48.45}  \\
					
		\hline
	\end{tabular}	}
\end{table}
\begin{table}[h]

	\scriptsize
	\renewcommand{\arraystretch}{1.1}
	\setlength{\tabcolsep}{2.5pt}
	\caption{Performance Comparison of state-of-the-art algorithms for the CUHK03 dataset on the `detected' setting. Proposed baseline S-CNN architecture outperforms all the previous state-of-the-art methods for CUHK03 dataset at Rank 1. The proposed variant of the S-CNN architecture with the gating function achieves the state-of-the-art results on CUHK03 benchmark dataset. In addition to the results at various ranks, we also provide the mean average precision to analyze the retrieval performance.}
	\label{cuhk03}
	\centering
	{
	\begin{tabular}{|c|c|c|c|c|}
		\hline
		\bfseries Method & \bfseries Rank 1 & \bfseries Rank 5 & \bfseries Rank 10 & \bfseries mAP 
		\\\hline\hline
		SDALF \cite{sdalf} & 4.9 & 21.0 & 31.7 & \_  \\

		ITML \cite{itml} & 5.14 & 17.7 & 28.3  & \_  \\

		LMNN \cite{Weinberger2009LMNN} & 6.25 & 18.7 & 29.0  & \_ \\

		eSDC \cite{zhao2013unsupervised} & 7.68 & 22.0 & 33.3  & \_  \\

		LDML \cite{isthatyou} & 10.9 & 32.3 & 46.7  & \_  \\

		KISSME \cite{kissmecvpr12} & 11.7 & 33.3 & 48.0  & \_   \\

		FPNN \cite{cuhk03} & 19.9 & 49.3 & 64.7  & \_ \\

		BoW \cite{market} & 23.0 & 45.0 & 55.7 & \_ \\
		BoW + HS \cite{market} & 24.3 & \_ & \_  & \_\\
 		ConvNet \cite{ejazdeep2015} 	& 45.0	& 75.3 & 83.4 & \_  \\
		LX \cite{lomo} & 46.3 &  78.9 & {88.6} & \_ \\
		MLAPG  \cite{mlapg} & 51.2 &  83.6 & {92.1} & \_ \\

		SS-SVM \cite{sssvm} & 51.2 & 80.8 & 89.6 & \_ \\
		SI-CI \cite{sicir} & 52.2 & 84.3 & 92.3 & \_ \\
		DNS \cite{dns} & 54.7 & {84.8} & {\bf 94.8} & \_ \\
		S-LSTM \cite{slstm} &{57.3} & {80.1}   &{88.3}    & {46.3}				 \\
		\hline
		{\bf \begin{tabular}{@{}c@{}} Ours - Baseline - S-CNN (SQ)\end{tabular}} & {58.1} & {79.2}  &{87.1} &{48.90}  \\
		{\bf \begin{tabular}{@{}c@{}} Ours - With Matching Gate (SQ)\end{tabular}} & {61.8}  & {80.9}  &{88.3} &{51.25} \\
		{\bf \begin{tabular}{@{}c@{}} Ours - Baseline - S-CNN (MQ)\end{tabular}} & {\bf 63.9} & {\bf 86.7}   &{92.6} & {\bf 55.57} \\
		{\bf \begin{tabular}{@{}c@{}} Ours - With Matching Gate (MQ)\end{tabular}} & {\bf 68.1} & {\bf 88.1}   & {94.6} & {\bf 58.84 }\\	
		\hline
	\end{tabular}	}
\end{table}

\begin{table}[!th]

	\scriptsize
	\renewcommand{\arraystretch}{1.2}
	\setlength{\tabcolsep}{2.5pt}
	\caption{Performance Comparison of state-of-the-art algorithms using an individual method for the VIPeR dataset. Proposed S-CNN framework outperforms several previous deep learning approaches for human re-identification \cite{ejazdeep2015,yi2014deep}. Our S-CNN with MG achieves promising results compared to other approaches.}
	\label{viper}
	\centering
	{
	\begin{tabular}{|c|c|c|c|c|c|}
		\hline
		\bfseries Method & \bfseries Rank 1 & \bfseries Rank 5 & \bfseries Rank 10 
		\\\hline\hline
		LFDA \cite{pedagadi2013local} & 24.1 & 51.2 & 67.1   \\

		eSDC \cite{zhao2013unsupervised} & 26.9 & 47.5 & 62.3    \\

		{\begin{tabular}{@{}c@{}}Mid-level \cite{zhao2014learning}\end{tabular}}  & 29.1 & 52.3 & 65.9     \\

		SVMML \cite{ZhenliShiyu_CVPR2013} & 29.4 & 63.3 & 76.3    \\

		VWCM \cite{zhang2014novel} & 30.7 & 63.0 & 76.0    \\

		SalMatch \cite{zhao2013person} & 30.2 & 52.3 & 65.5    \\

		QAF \cite{queryadapt} & 30.2 & 51.6 & 62.4  \\
		
		DML \cite{yi2014deep} & 28.2 & 59.3 & 73.5   \\

		CMWCE \cite{yangcolor2014} & 37.6 & 68.1 & {81.3}   \\

		{\begin{tabular}{@{}c@{}}SCNCD \cite{salientcolorECCV14} \end{tabular}} &  37.8 & 68.5 & 81.2   \\

		LX \cite{lomo} & 40.0 & 68.1 & 80.5   \\
		PRCSL \cite{prcsl} & 34.8 & 68.7 &	{82.3} \\
		MLAPG  \cite{mlapg} & 40.7	&	69.9 &	{82.3}\\
		MT-LORAE \cite{mtlorae} & 42.3 & {\bf72.2} & 81.6 \\
		Semantic \cite{transfsem} & 41.6 & 71.9  & {\bf 86.2}\\
		
		S-LSTM \cite{slstm} & {42.4}   & {68.7}  & {79.4}   \\

		SS-SVM \cite{sssvm} & 42.7 & \_	&	84.3 \\
		
		HGD \cite{hgdreid} & 49.7 & 79.7	&	88.7 \\
		DNS \cite{dns} & 51.7 & 82.1 & 90.5 \\		
		SCSP \cite{scsp} & {\bf 53.5} & {\bf 82.6} & {\bf 91.5} \\
		
		\hline
		ConvNet \cite{ejazdeep2015} 	& 34.8	& 63.7 & 75.8   \\
		SI-CI \cite{sicir} & 35.8 & 67.4 & 83.5 \\
		DGDropout \cite{domainguided} & 38.6 & \_ & \_ \\
		MCP-CNN \cite{mcpbc} & 47.8 & 74.7 & 84.8 \\
		\hline		

		{\bf \begin{tabular}{@{}c@{}} Ours - Baseline - S-CNN \end{tabular}} & {36.2} & {65.1}   &{76.3} \\
		{\bf \begin{tabular}{@{}c@{}} Ours - With Matching Gate\end{tabular}} & {37.8} & {66.9}   &{77.4}\\	
		\hline
	\end{tabular}	}
\end{table}

\subsection{Results and Discussion}The results for the Market-1501, CUHK03 and VIPeR datasets are given in Tables \ref{market}, \ref{cuhk03} and \ref{viper} respectively. The proposed baseline S-CNN architecture outperforms all the existing approaches for human re-identification for Market-1501 and CUHK03 datasets at Rank 1. We believe that the baseline S-CNN architecture sets a strong baseline for comparison of supervised techniques in future works for both datasets. However, for VIPeR dataset, even though our baseline S-CNN does not achieve the best results, it outperforms several other CNN based architectures \cite{ejazdeep2015,yi2014deep,sicir}. Our final architecture with the MG improves over the baseline architecture by a margin of $4.2\%$ and $1.6\%$ at Rank 1 for CUHK03 and VIPeR datasets respectively. For Market-1501 dataset, our approach outperforms the baseline by a margin of $3.56\%$ at Rank 1 for single query (SQ) setting and $3.12\%$ at Rank 1 for multi query (MQ) setting.
\begin{figure}[!t]
\centering
\includegraphics[width=0.8\linewidth]{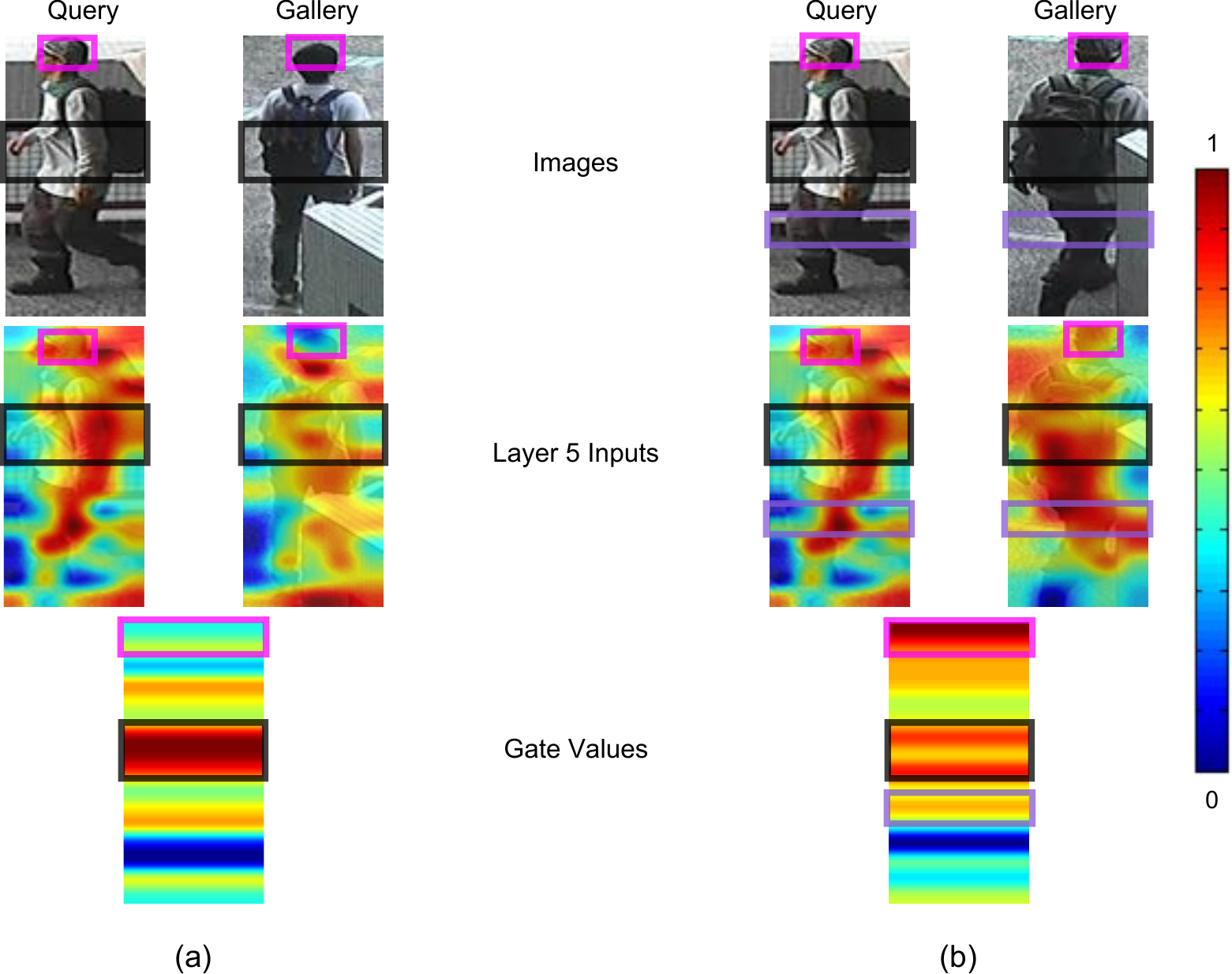}
\caption{{\bf Gate Visualization: }(a) Query paired with its hard-negative (b) Query paired with its positive. Middle row shows the layer 5 input values of all the 4 images and last row shows the corresponding gate values obtained for both pairs. Boxes of same color indicates corresponding regions in the images. {\bf Best viewed in color}}
\label{fig:viz}
\end{figure}

For multi-camera networks, the mean average precision is a better measure for performance compared to the Rank - 1 accuracy \cite{market} as it signifies how many of the correct matches are retrieved from various camera views. Therefore, compared to the improvement in Rank 1 accuracy, the mean average precision which indicates the retrieval accuracy may be more interesting for real-world applications with camera networks. Even though the mean average precision is not particularly important for CUHK03 dataset as it contains only two views, we report the mAP to compare the retrieval results of the proposed final architecture with the baseline S-CNN architecture. It can be seen that our final architecture with MG outperforms the mean average precision obtained by the baseline S-CNN by a margin of $3.32\%$, $3.06\%$ and $3.27\%$ for Market-1501-Single Query, Market-1501-Multi Query and CUHK03 datasets respectively.

The visualization of the gating mechanism in the proposed matching gate is shown in Figure \ref{fig:viz}. Figure \ref{fig:viz} (a) shows a query image and a hard negative image (example shown in Figure \ref{fig:example} (b)). The middle row shows the average feature activations at the output of the $4^{th}$ convolutional block which is the input to the proposed gating function and the third row shows the obtained gate values using the proposed gating function. It can be seen that for the first few rows where the subject in the query is wearing a hat, the gate activations are low indicating lower similarity where as for a few middle rows, the gate activations are high indicating higher similarity. In Figure \ref{fig:viz} (b), we show the image paired with its true positive, the layer 5 inputs and the gate values. It can be seen that for majority of the patches, the gate values are high indicating high similarity between the image patches. This indicates that the gating function can efficiently extract relevant common information from the feature maps of both the images and boost them.

\section{Conclusion and Future Works}
We have proposed a baseline siamese CNN and a learnable Matching Gate function for siamese CNN that can vary the network behavior during training and testing for the task of human re-identification. The Matching Gate can compare the local features along a horizontal stripe for an input image pair during run-time and adaptively boost local features for enhancing the discriminative capability of the propagated features. The gating function is also designed to be a differentiable one with learnable parameters for adjusting the variance of the gate values as well as for summarizing the horizontal stripe features. This is essential for adjusting the amount of filtering at each stage of the network as well as to facilitate end-to-end learning of deep networks. We have conducted experiments on the Market-1501 dataset, the CUHK03 dataset and the VIPeR dataset to evaluate how run-time feature selection can enable the network to learn more discriminative features for extracting meaningful similarity information for an input pair. The introduction fo the gating function in between convolutional layers results in significant improvement of performance over the baseline S-CNN. Our S-CNN model with the matching gate achieves promising results compared to the state-of-the-art algorithms on the above datasets.

\subsubsection*{Acknowledgments: }The research is supported by Singapore Ministry of Education (MOE) Tier 2 ARC28/14, and Singapore A*STAR Science and Engineering Research Council PSF1321202099. 

This research was carried out at the Rapid-Rich Object Search (ROSE) Lab at Nanyang Technological University. The ROSE Lab is supported by the National Research Foundation, Singapore, under its Interactive Digital Media (IDM) Strategic Research Programme. 

We thank NVIDIA Corporation for their generous GPU donation to carry out this research.

\bibliographystyle{splncs03}
\bibliography{egbib}
\end{document}